\pdfoutput=1
\documentclass[10pt,twocolumn,letterpaper]{article}

\usepackage[letterpaper,margin=0.75in]{geometry}
\usepackage{times}
\usepackage{xcolor}
\usepackage{caption}
\usepackage[numbers,sort&compress]{natbib}
\usepackage{url}
\usepackage[breaklinks,colorlinks=true,allcolors=blue]{hyperref}

\usepackage{amsmath,amssymb}
\usepackage{booktabs}
\usepackage{array}
\usepackage{tabularx}
\usepackage{graphicx}
\usepackage{enumitem}
\usepackage{multirow}
\usepackage{algorithm}
\usepackage{algorithmic}
\title{Interactive Person Retrieval with On-the-Fly Visual Feedback}
\author{%
  Lyucheng Qian$^{1}$ \quad John Yuehan Zhang$^{2}$ \quad Pingyu Wang$^{1}$\thanks{Corresponding author}\\
  $^{1}$Sichuan University \quad $^{2}$University of California, Berkeley
}
\date{}

\begin{document}
\maketitle

\begin{abstract}
Traditional text-based person retrieval (TPR) assumes that a witness provides a complete description at once, whereas interactive person retrieval (Inter-PR) refines a partial description through question--answer exchanges. In this setting, the questioner inspects the retrieved candidates to decide what evidence to request next, while the answerer sees only the witness-side memory and cannot report candidate-specific visual differences. Existing methods mainly learn offline question orderings, leaving this visual-feedback loop underexplored. To address this limitation, we introduce visual-feedback interactive person retrieval (\textbf{VFI-PR}), a new interactive retrieval paradigm in which retrieved candidates produce visual feedback that becomes evidence for subsequent questioning and re-ranking. To support training and evaluation in this setting, we construct \textbf{VFI-PEDES}, a dataset and evaluation protocol for visual-feedback interaction, and propose \textbf{VFI-ReID}, a framework that learns to use visual feedback during interactive person retrieval. Through the Dynamic Accumulated-evidence Retriever (DAR) and retrieval-aware question optimization, VFI-ReID adapts retrieval and question generation to evolving evidence states, turning visual feedback into complementary evidence for subsequent retrieval. Experiments show that VFI-ReID consistently improves retrieval across VFI-PR interaction stages, demonstrating the utility of visual feedback for interactive person retrieval and outperforming existing Inter-PR baselines.
\begin{figure}[t]
    \centering
    \includegraphics[width=\columnwidth]{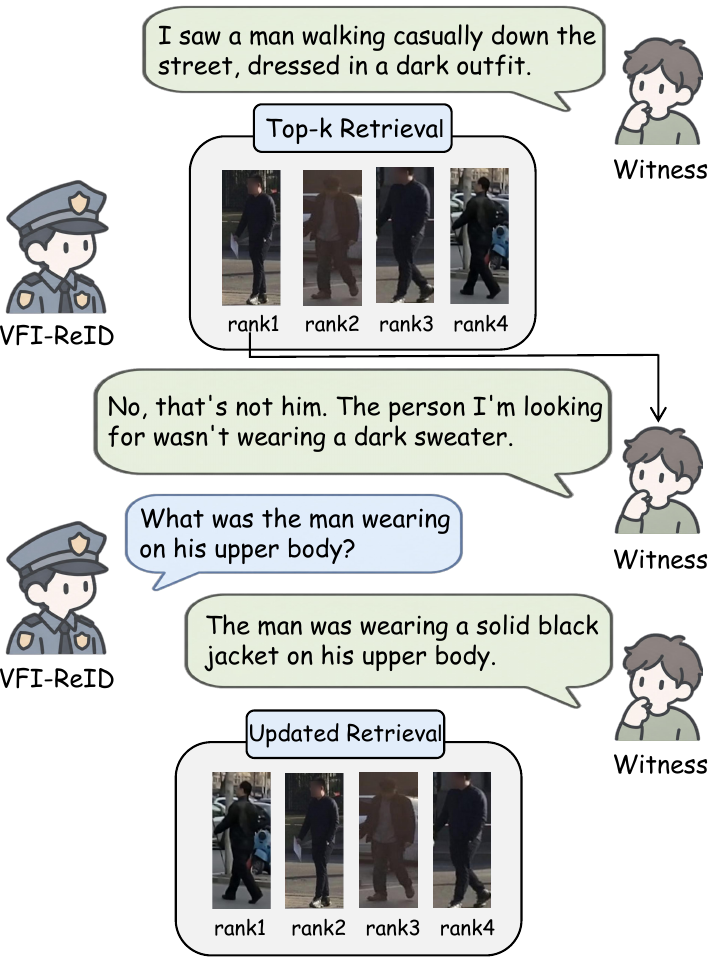}
\caption{Illustration of VFI-PR with VFI-ReID. Retrieved candidates provide visual corrections, after which the questioner acquires complementary evidence and the ranking is updated.}
    \label{fig:teaser}
\end{figure}
\end{abstract}

\section{Introduction}
\label{sec:intro}

Person retrieval (PR) aims to retrieve images of the same person from a gallery given a query image or a natural-language description \cite{li2014deepreid,he2021transreid}. Traditional text-based person retrieval (TPR) is attractive when the target image is unavailable, but a short description is often incomplete, ambiguous, and biased toward attributes that are easy to verbalize. Existing TPR systems therefore learn a cross-modal similarity function from a single text query, even though the evidence needed to distinguish a target from visually similar people is rarely available in one pass.

To acquire such additional evidence, interactive person retrieval (Inter-PR) allows a system to ask questions and incorporate the witness's answers into the query. ChatIR uses a language-model questioner to conduct multi-turn dialogue and clarify an initial image description \cite{levy2023chatting}. PlugIR reformulates dialogue context for black-box retrievers, grounds question generation in the current retrieval context, and filters redundant questions \cite{lee2024plugir}. LLaVA-ReID introduces a multimodal questioner with selective visual context and looking-forward supervision for learning retrieval-oriented question orders \cite{lu2025llavareid}. Although these methods make retrieval iterative, their answerer remains unaware of the retrieved candidates. Consequently, the answerer cannot describe how the current candidates differ from the target, so the dialogue lacks a direct visual feedback channel.

More recently, MNEMO introduced Multimodal Interactive Person Retrieval (MInter-PR), a multimodal conversational retrieval setting in which users describe visual differences between a target and retrieved candidates \cite{bai2026interactive}. MNEMO shows that candidate images can become a source of new evidence rather than merely the output of retrieval. However, it has three limitations for a real-time feedback loop. First, it does not produce visual feedback after every interaction round, so feedback is not continuously interleaved with questioning and retrieval. Second, its retriever is trained on one fully concatenated dialogue text rather than on the variable-length intermediate states encountered during an evolving dialogue. Third, its dataset construction uses the two candidate images selected for visual feedback with prior knowledge that neither depicts the target, whereas such target-exclusion information is unavailable in real deployment. These differences make the original setting difficult to apply directly when candidate identities are unknown and feedback must be generated online.

Motivated by this candidate-feedback setting, we formulate visual-feedback interactive person retrieval (VFI-PR). The central design principle is that feedback is evidence tied to the current retrieval state, not a post-hoc annotation attached only after the conversation. At each round, the answerer first describes a visual difference grounded in the current candidate; the questioner then asks a question and the answerer responds. These three pieces are incorporated into the accumulated evidence in the order feedback, question, and answer, after which the retriever updates the gallery ranking. This ordering makes the next question condition on both what the witness has said and what the current retrieval has failed to explain.

To facilitate the study of VFI-PR, we construct the VFI-PEDES dataset and develop the Visual-Feedback Interactive Re-Identification (VFI-ReID) framework. VFI-PEDES reuses the image identities and initial descriptions of Interactive-PEDES, but adds visual-difference feedback and records the resulting evidence states. On top of this dataset, VFI-ReID applies the Dynamic Accumulated-evidence Retriever (DAR) and trains the multimodal questioner in two stages: supervised fine-tuning (SFT) serves as a warm-up, followed by retrieval-aware reinforcement learning (RL). This arrangement first adapts the retriever and questioner to the visual-feedback interaction, then optimizes questions according to the ranking changes produced by the complete question--answer--retrieval loop.

In summary, this paper makes the following contributions:
\begin{itemize}[leftmargin=*,itemsep=1pt,topsep=2pt]
    \item We formulate VFI-PR as a closed-loop evidence-acquisition problem in which retrieved candidates generate feedback that conditions later questions.
    \item We construct VFI-PEDES, a dataset with visual-difference feedback, explicit feedback-selection rules, and accumulated evidence states for reproducible VFI-PR training and evaluation.
    \item We develop a two-stage training strategy for the multimodal questioner, combining supervised warm-up with retrieval-aware policy optimization; DAR adapts the retriever to the accumulated evidence states used by this strategy.
    \item We show experimentally that VFI-ReID improves retrieval throughout VFI-PR interactions, and analyze how visual feedback, retriever adaptation, supervised warm-up, and policy optimization contribute to this improvement.
\end{itemize}

\begin{figure*}[t]
    \centering
    \includegraphics[width=\textwidth]{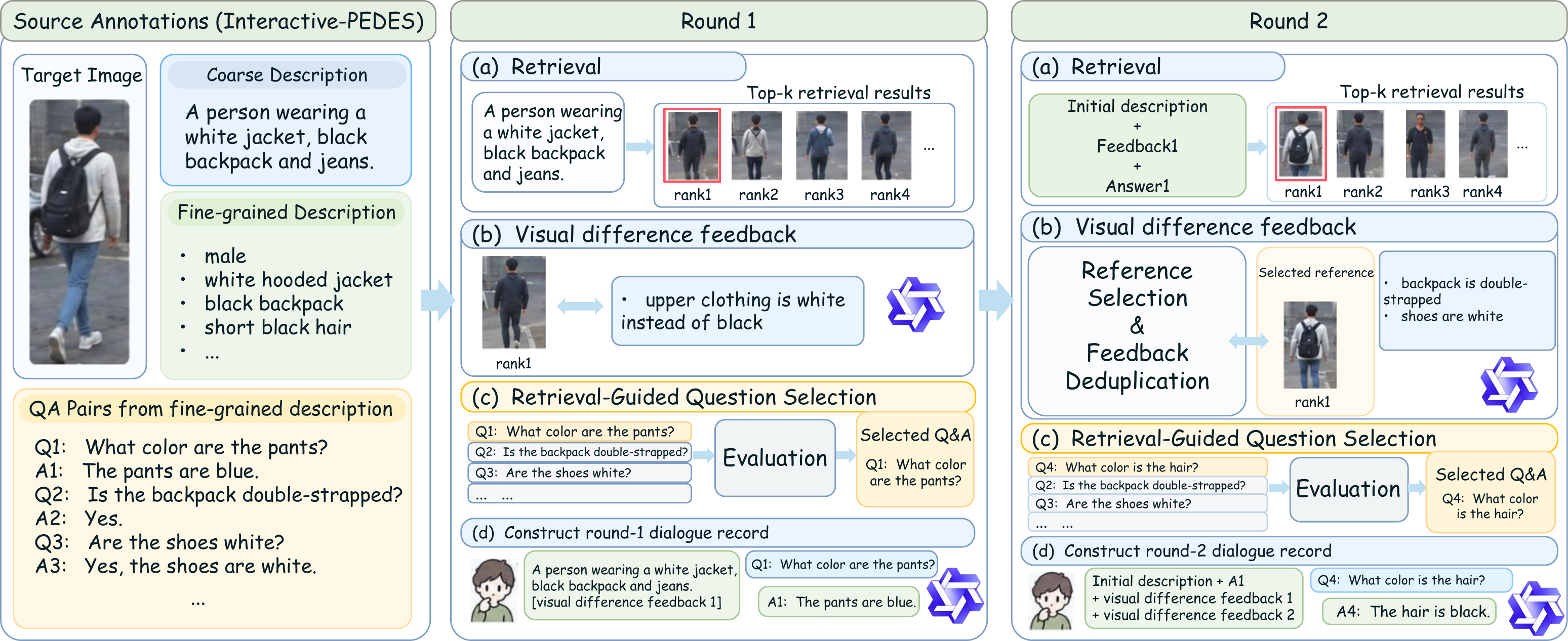}
\caption{Overview of the VFI-PEDES construction. Visual-difference feedback is generated from retrieved candidates, and Retrieval-Guided Question Selection uses simulated retrieval gains to select question--answer pairs and construct accumulated evidence states.}
    \label{fig:pipeline}
\end{figure*}

\section{Related Work}
\label{sec:related}

\subsection{Text-based Person Retrieval}
Traditional person re-identification matches a query image against a gallery, whereas text-based person retrieval (TPR) performs the same search from a natural-language description. CUHK-PEDES established the first large-scale benchmark for cross-modal person retrieval \cite{li2017cuhkpedes}, while ICFG-PEDES introduced a separate identity-centric and fine-grained benchmark \cite{ding2021ssan}. Subsequent TPR methods improved image--text alignment through part-aware modeling, semantic relations, attribute supervision, and uncertainty-aware representations \cite{wang2020vitaa,chen2021tipcb,zhu2021dssl,jiang2023cross,yan2023cfine,bai2023rasa,yang2023aptm,li2024aul,qin2024rde}. CLIP-style pretraining and multimodal foundation models further strengthen this matching process \cite{radford2021clip,li2021albef,li2022blip,li2024llavaonevision}. Interactive-PEDES extends the setting by decomposing descriptions into multi-round question--answer interactions \cite{lu2025llavareid}.

Despite these advances, TPR methods generally treat the description as a completed, fixed query: they improve alignment between clothing, appearance, and contextual cues, but do not decide which missing attribute should be requested or adapt ranking as evidence accumulates. VFI-PR addresses this limitation by making variable-length evidence composition part of retrieval itself: the retriever must continually consume an initial description together with visual-difference feedback and answers as the interaction proceeds.

\subsection{Interactive person retrieval}
Interactive retrieval treats search as sequential evidence acquisition rather than a one-shot match \cite{das2017visualdialog,devries2017guesswhat}. ChatIR and PlugIR study conversational image retrieval and query refinement, while SimRV uses question--answer interaction for video retrieval \cite{levy2023chatting,lee2024plugir,liang2023simple}. ChatPR extends this direction to person retrieval with dialogue-refined cross-modal alignment \cite{bai2025chatpr}, and LLaVA-ReID introduces Interactive-PEDES together with a multimodal questioner that uses candidate images and retrieval-based supervision to order questions offline \cite{lu2025llavareid}. Human-centered work has also explored multimodal large language model (MLLM) assistance for retrieval interactions, but uses the model as an auxiliary component rather than learning a complete interactive question policy \cite{qin2025human}. MNEMO further introduces multimodal conversational retrieval with visual-difference feedback, allowing retrieved candidates to contribute evidence to the conversation \cite{bai2026interactive}.

These studies establish targeted questioning and multimodal conversational retrieval, but they do not provide a fully interleaved feedback--question--answer loop in which both the retriever and questioner adapt to every intermediate state. VFI-PR builds on this foundation by generating visual-difference feedback throughout the interaction, adapting the retriever to variable-length accumulated evidence, and training the questioner to use the resulting retrieval states.

\section{Benchmark}
\label{sec:dataset}

In this section, we introduce VFI-PEDES as the benchmark and evaluation protocol for VFI-PR. Figure~\ref{fig:pipeline} summarizes how VFI-PEDES is constructed. We first define VFI-PR and then describe the construction process.

\subsection{Task definition}
VFI-PR starts from an initial description $x_0$, a target identity $y$, and an image gallery $\mathcal{G}$. The interaction uses candidate sets of size $k$ and an evidence-acquisition budget of $T$ rounds. Because the answerer cannot inspect the hidden target image in a real interaction, the fine-grained description serves as its memory. At interaction round $t$, the cross-modal retriever uses the accumulated evidence $m_t$ to produce the ranked candidate set $C_t$. The selected candidate is shown to the answerer, who provides a short visual-difference feedback sentence $f_t$ describing the difference between the target and the selected candidate. The questioner then asks about an unresolved attribute using $m_t$ and $C_t$, and the answerer provides the corresponding information from the memory. The question remains in the dialogue context, whereas the retrieval evidence is updated by appending the feedback and the cleaned answer in that order:
\begin{equation}
    m_{t+1}=m_t\oplus f_t\oplus \tilde{a}_t,
    \label{eq:evidence-update}
\end{equation}
where $\oplus$ denotes ordered text concatenation and $\tilde{a}_t$ is the answer after the shared cleaning rule.

\subsection{Data construction}
\label{sec:finegrained-retriever}
VFI-PEDES starts from the images, identities, initial descriptions, and question--answer annotations of Interactive-PEDES. We preserve the original answers and add feedback generated from the current candidate state. During construction, we use each fine-grained description as the answerer's memory, reflecting the realistic setting in which a witness can describe the target but cannot provide its image. Given this memory and the selected candidate image, the answerer produces one complete, ordinary-language visual-difference feedback sentence describing the difference between them. The answerer does not see the target image, so the feedback remains in the same textual channel as the accumulated evidence without leaking the target image.

The first feedback is generated from the initial Rank-1 candidate. After the first feedback, question, and answer are incorporated into the accumulated evidence, the retriever produces a new ranked candidate set for the next feedback. We use the \textbf{Reference Selection and Feedback Deduplication} procedure in Algorithm~\ref{alg:ivfs} to select this reference: if the new Rank-1 has the same identity as the first reference, the highest-ranked candidate with a different identity is selected; if all candidates in the current set share that identity, the new Rank-1 is retained as a fallback. The procedure uses only gallery-side identity metadata, since standard ReID galleries can group images by indexed identity even when the query identity is unknown. It compares retrieved candidates only, never the target identity or image, and therefore enforces candidate diversity without target-aware information. If the resulting feedback duplicates the preceding feedback, it is skipped rather than appended as repeated evidence.

\begin{algorithm}[!t]
\caption{The process of reference selection and feedback deduplication}
\label{alg:ivfs}
{\raggedright\noindent\textbf{Input:} Selected candidate $c_1$, current ranked Top-$k$ candidate set $C_2$, and preceding feedback $f_1$.\par}
\begin{algorithmic}[1]
\STATE $c_2 \leftarrow C_2[1]$
\IF{$\mathrm{ID}(c_2)=\mathrm{ID}(c_1)$}
    \STATE $C_{\mathrm{diff}}\leftarrow\{c\in C_2:\mathrm{ID}(c)\neq\mathrm{ID}(c_1)\}$
    \IF{$C_{\mathrm{diff}}\neq\varnothing$}
        \STATE $c_2\leftarrow$ highest-ranked candidate in $C_{\mathrm{diff}}$
    \ENDIF
\ENDIF
\IF{$\mathrm{ID}(c)=\mathrm{ID}(c_1)$ for all $c\in C_2$}
    \STATE $c_2\leftarrow C_2[1]$ \COMMENT{fallback}
\ENDIF
\STATE Generate $f_2$ from the memory and $c_2$
\IF{$f_2=f_1$}
    \STATE $f_2\leftarrow\varnothing$ \COMMENT{skip duplicate feedback}
\ENDIF
\end{algorithmic}
{\raggedright\noindent\textbf{Output:} Next candidate $c_2$ and feedback $f_2$ or $\varnothing$.\par}
\end{algorithm}

The question labels and retriever states are constructed with Retrieval-Guided Question Selection. To make this selection retrieval-aware, we first train a fine-grained-description retriever by aligning each image with its corresponding fine-grained description through image--text contrastive learning. It provides the ranking signal for measuring how much each candidate question--answer pair improves the target rank and serves as the baseline retriever in the ablation study. Given the current evidence $m_t$, we simulate every answerable question--answer pair by appending the question and its cleaned answer, then use this retriever to measure the resulting target rank. Subject to the answerability and duplication rules, the optimal pair is selected as
\begin{equation}
 (q_t^\star,a_t^\star)
 = \arg\max_{(q,a)\in\mathcal{P}_t}
 \bigl[r_t-r_t(q,a)\bigr],
 \label{eq:rgqs}
\end{equation}
where $\mathcal{P}_t$ is the set of answerable, non-duplicate question--answer pairs, $r_t$ is the current target rank, and $r_t(q,a)$ is the target rank after the simulated pair is appended. Repeating this step after each selected pair produces the question--answer trajectories and variable-length evidence states used by VFI-PEDES.

Finally, we construct the VFI-PEDES dataset. Its training split contains 47,376 images from 11,543 identities, while its test split contains 7,373 images from 1,508 identities. Each record supports two interaction rounds and exposes five accumulated-text states, consisting of the initial description and the feedback and answer states produced during the interaction. Further construction details are provided in the Supplementary Material, Sec. 1.1.

\begin{figure*}[!t]
    \centering
    \includegraphics[width=\textwidth]{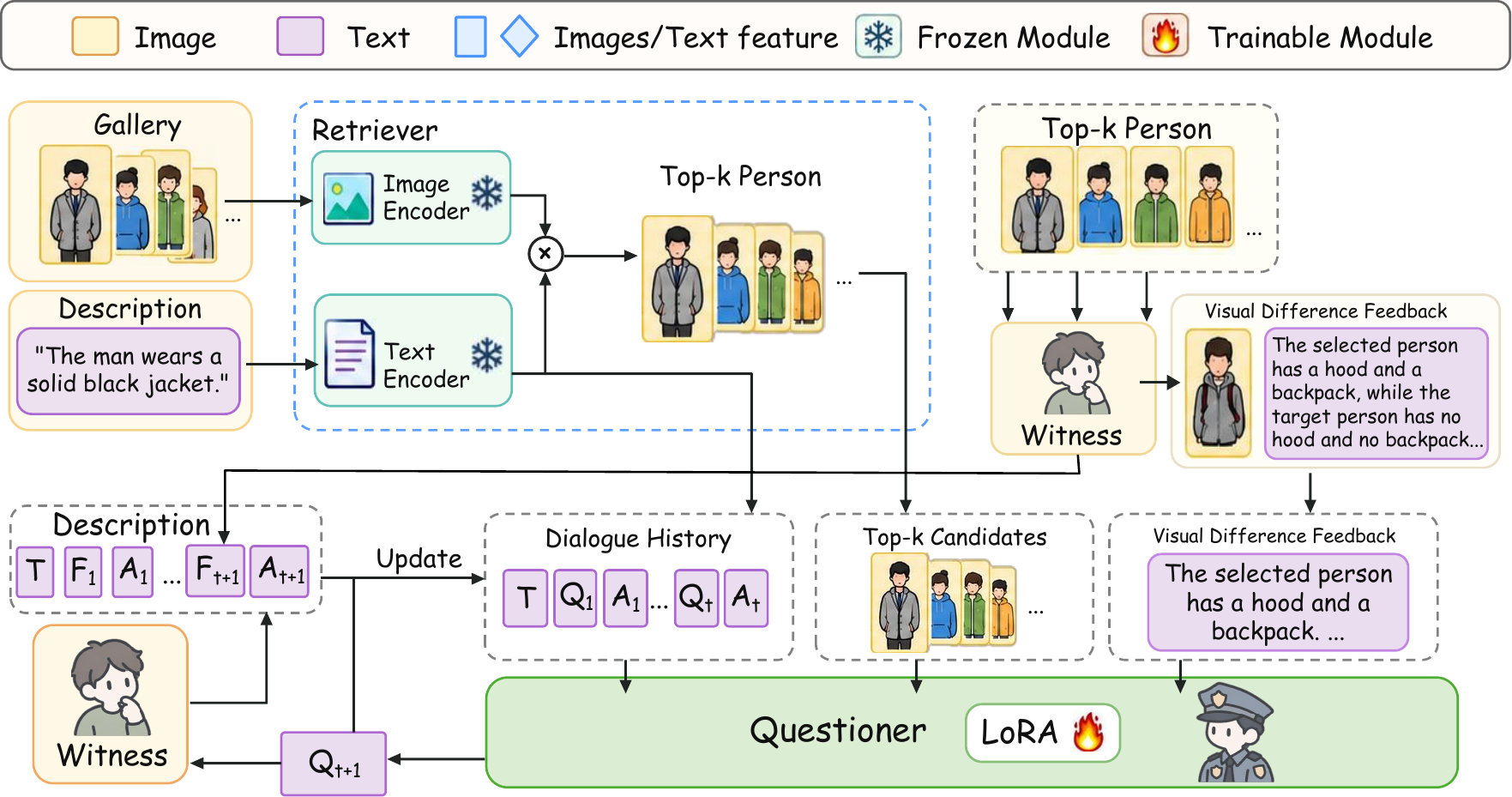}
    \caption{Overview of the VFI-ReID framework for VFI-PR. The Retriever constructs the current Top-$k$ candidates from image--text features. The Witness produces visual-difference feedback and answers the Questioner's question. The Questioner uses the memory, dialogue history, candidates, and available feedback to acquire the next answer and update the retrieval text.}
    \label{fig:method-pipeline}
\end{figure*}

\section{Method}
\label{sec:method}

\subsection{Visual-feedback interactive ReID framework}

The VFI-ReID framework consists of three main components: a Retriever, a Questioner, and an Answerer. Given the current accumulated evidence and a gallery $\mathcal{G}$, the Retriever produces the ranked candidates for the current interaction state. The Questioner uses this state to ask for the next unresolved identity cue, while the Answerer supplies either visual feedback or the witness-side response. The accumulated evidence is then updated and the Retriever recomputes the gallery ranking. This design extends the conventional Inter-PR loop with an explicit feedback channel while preserving the question--answer interaction used by existing systems.

The Retriever consumes the current accumulated evidence together with the gallery images and produces the ranking for the current interaction state. Its Top-$k$ candidates provide the visual context used by the Questioner and determine which candidate is selected for feedback. Unlike a static text retriever, it must handle evidence states that grow and change as feedback, questions, and answers enter the interaction. The evidence-adaptation procedure is described in Sec.~\ref{sec:dar}.

The Questioner reads the initial description, the dialogue history, the current Top-$k$ candidates, and available visual-difference feedback before asking about an unresolved identity cue. Its question is selected for the evidence that its answer is expected to contribute to the next retrieval update. We first learn this behavior from retrieval-guided question labels and then refine it with retrieval-aware policy optimization, as detailed in Sec.~\ref{sec:questioner}.

The Answerer supplies the witness-side response to the selected question from the available memory. It also provides the visual-difference feedback used by VFI-PEDES: given the memory and a selected candidate, it describes a feature of the target that is not visible on that candidate. The answerer does not see the target image, so both kinds of answer remain grounded in the witness-side textual channel described in Sec.~\ref{sec:dataset}.

\subsection{Dynamic accumulated-evidence retriever}
\label{sec:dar}

The retriever in a conventional TPR pipeline is trained on a fixed description distribution. That distribution does not contain the feedback--answer compositions encountered in VFI-PEDES. As described in Sec.~\ref{sec:finegrained-retriever}, we first train the fine-grained-description retriever and use it for Retrieval-Guided Question Selection, obtaining the question--answer trajectories and variable-length evidence states. To match this evolving input, DAR adapts the cross-modal retriever to these dynamic-length states. After DAR is trained, the Retriever is frozen during Questioner training.

For each target identity, DAR creates a state sequence
\begin{equation}
    \mathcal{M}=\{m_0, m_0\oplus f_1, m_0\oplus f_1\oplus a_1,\ldots\}.
\end{equation}
Each state is paired with the same target image and sampled non-matching gallery images. We follow the IRRA objective \cite{jiang2023cross}, whose image--text alignment component pulls matched image--text pairs together and separates mismatched pairs, while its cross-modal matching component matches the image--text similarity distributions so that the target pair receives higher compatibility than other gallery pairs. We apply these two components to every accumulated-evidence state:
\begin{equation}
    \mathcal{L}_{\mathrm{DAR}}=\mathcal{L}_{\mathrm{align}}(m_t,I_y)+\lambda\mathcal{L}_{\mathrm{match}}(m_t,I_y),
\end{equation}
where $I_y$ is the target image, $\mathcal{L}_{\mathrm{align}}$ and $\mathcal{L}_{\mathrm{match}}$ are the alignment and similarity-distribution matching components adopted from IRRA, and $\lambda$ balances them. Sampling states with different lengths prevents the model from treating feedback as an out-of-distribution suffix and teaches it to use positive, negative, and corrective evidence consistently. At VFI-PR test time, DAR receives only the current accumulated text and the gallery images.

\subsection{Evidence-acquisition questioner}
\label{sec:questioner}

At each interaction step, the questioner reads the initial description, the dialogue history, the current Top-$k$ candidates, and available visual-difference feedback before asking a free-form question about an unresolved identity cue. The resulting answer is incorporated into the retrieval text so that the retriever can update the ranking. To make the questions both answerable by the witness and useful for retrieval, we train the questioner in two stages: a supervised warm-up first teaches the desired question behavior from retrieval-guided labels, and retrieval-aware policy optimization further aligns question selection with retrieval outcomes.

\subsubsection{Supervised warm-up}
\label{sec:sft}

SFT provides the supervised warm-up stage by converting the offline question labels selected in Sec.~\ref{sec:dar} into an explicit question-generation task. Following Figure~\ref{fig:method-pipeline}, the questioner reads the initial description, the dialogue history containing previous question--answer exchanges, the ranked Top-$k$ candidates from DAR, and available visual-difference feedback. For sample $i$, this context is written as
\begin{equation}
    x_i=\big[d_i,\;h_i,\;C_i,\;f_i\big],
\end{equation}
where $d_i$ is the initial witness description, $h_i$ is the dialogue history containing previous question--answer exchanges, $C_i$ is the current ranked candidate set, and $f_i$ is the available visual-difference feedback; $f_i$ is empty when no feedback is available at that state. The desired output is a single free-form question that asks for an unresolved identity attribute and can be answered by the witness. The selected question $q_i^*$ is the label associated with sample $i$; its answer is used during dataset construction to determine the retrieval-aware label, but the answer itself is not part of the questioner's input. This distinction lets SFT learn from retrieval consequences while preserving the causal test-time protocol.

The questioner is initialized from a multimodal language model and fine-tuned with QLoRA \cite{dettmers2023qlora}. The supervised objective for the selected question is
\begin{equation}
    \mathcal{L}_{\mathrm{SFT}}(\theta)
    =-\frac{1}{|q_i^*|}
    \sum_{\ell=1}^{|q_i^*|}
    \log\pi_\theta(q_{i,\ell}^*\mid x_i,q_{i,<\ell}^*).
\end{equation}
Here, $x_i$ is the four-part context for sample $i$, $q_i^*$ is its selected question, and $q_{i,<\ell}^*$ is the prefix of the current question before token $\ell$. Thus, the label is the Retrieval-Guided Question Selection output. Only feedback available before the question is included; if a feedback step is skipped by the VFI-PEDES rule, its field is empty rather than duplicated.

The supervised warm-up provides a stable question-generation policy, but it also exposes a limitation of imitation: the SFT questioner may repeat an attribute already established by visual feedback instead of exploring another unresolved cue. This repetition consumes the current interaction step without eliciting new evidence, even when other attributes could further distinguish the target from the candidate. We therefore use retrieval-aware policy optimization to favor questions whose answers improve retrieval.

\subsubsection{Retrieval-aware reinforcement learning}
\label{sec:rl}

To construct the RL pool, we start from the VFI-PEDES training states and retain cases whose target is not already at Rank-1 and that have at least one answerable, non-duplicate question--answer candidate. This focuses policy learning on states where another question can still change the retrieval outcome, rather than spending capacity on already solved cases. We stratify the retained states by evidence length so that shorter and longer interaction states are represented throughout training, and preserve the source-dataset proportions of VFI-PEDES before random sampling so that neither constituent dataset dominates the pool. We sample 3K such states for the main model, while 1K and 5K pools are used in the scaling study. Further details of the RL dataset are provided in the Supplementary Material, Sec. 2.2.

We then refine the SFT policy with Group Relative Policy Optimization (GRPO) \cite{shao2024deepseekmath}. The retriever and answerer remain fixed, and only the questioner's trainable parameters are updated. At each fixed state, the policy samples a group of questions and evaluates them independently under the same witness, candidate set, and retrieval environment. A validity gate first checks whether a question is answerable under the witness protocol. A valid question is sent to the fixed answerer; its response is cleaned with the shared answer-cleaning rule, appended to the current retrieval text, and passed to DAR for re-ranking. Invalid questions do not enter the answerer or retriever, but remain in the sampled group so that the policy is penalized for producing them. The reward combines the realized retrieval gain with a Rank-1 boundary bonus and two penalties. The retrieval term measures the reciprocal-rank improvement rather than the surface form of the question. Let $r^{\mathrm{before}}$ and $r^{\mathrm{after}}$ be the target ranks before and after the complete evidence update:
\begin{equation}
    \Delta\mathrm{RRG}=
    \frac{1}{r^{\mathrm{after}}}
    -\frac{1}{r^{\mathrm{before}}}.
\end{equation}
\begin{table*}[t]
\centering
\caption{VFI-PR retrieval performance before interaction and after each complete interaction round. Best log Rank Integral (BRI) \citep{lee2024plugir} is lower when retrieval improves earlier. Boldface indicates the best result, and underlining indicates the second-best result.}
\label{tab:legacy}
\small
\begin{tabularx}{\textwidth}{>{\centering\arraybackslash}p{1.2in}>{\centering\arraybackslash}p{0.72in}|*{4}{>{\centering\arraybackslash}X}|*{4}{>{\centering\arraybackslash}X}|>{\centering\arraybackslash}X}
\toprule
\multirow{2}{*}{Method} & \multirow{2}{*}{Venue} & \multicolumn{4}{c|}{Round 1} & \multicolumn{4}{c|}{Round 2} & \multirow{2}{*}{BRI$\downarrow$} \\
& & R@1 & R@5 & R@10 & mAP & R@1 & R@5 & R@10 & mAP & \\
\midrule
Initial & - & 62.05 & 82.87 & 89.72 & 47.45 & 62.05 & 82.87 & 89.72 & 47.45 & - \\
ChatIR~\citep{levy2023chatting} & NeurIPS'23 & 73.51 & 90.34 & 94.58 & 54.23 & 76.25 & 92.15 & 95.88 & 56.31 & 0.455 \\
PlugIR~\citep{lee2024plugir} & ACL'24 & 73.27 & \underline{90.51} & 94.87 & \underline{54.24} & 76.18 & 91.94 & 95.62 & 56.08 & 0.464 \\
SimRV~\citep{liang2023simple} & ICCV'23 & \underline{73.92} & 90.33 & \underline{94.96} & 53.64 & \underline{76.46} & \underline{92.28} & \underline{95.97} & \underline{56.59} & \underline{0.446} \\
LLaVA-ReID~\citep{lu2025llavareid} & ICML'25 & 72.25 & 89.49 & 94.18 & 53.39 & 73.95 & 90.55 & 94.71 & 54.76 & 0.469 \\
VFI-ReID~(Ours) & - & \textbf{76.43} & \textbf{92.11} & \textbf{95.69} & \textbf{55.63} & \textbf{80.14} & \textbf{93.80} & \textbf{96.66} & \textbf{57.77} & \textbf{0.385} \\
\bottomrule
\end{tabularx}
\end{table*}

An invalid-question penalty prevents shortcuts such as asking for the candidate index, requesting a ranking, filling in the answer itself, dumping the witness memory, or generating a question that cannot be answered by the witness. We additionally give a fixed Rank-1 bonus $b=0.35$ when a question moves the target from a non-Rank-1 position into Rank-1, and set $b=0$ otherwise, so that successful target acquisition is distinguished from smaller rank improvements. The reward is therefore
\begin{equation}
 R(q)=
 \begin{cases}
    \Delta\mathrm{RRG}+b-\lambda_{\mathrm{rep}}\mathbf{1}[\mathrm{group\ repeat}],
     & q\in\mathcal{V},\\
 -\lambda_{\mathrm{inv}}, & q\notin\mathcal{V},
 \end{cases}
\end{equation}
where $\mathcal{V}$ is the set of valid questions. We set $\lambda_{\mathrm{inv}}=0.2$ for invalid questions, while $\lambda_{\mathrm{rep}}=0.02$ is applied to group repeats among questions sampled from the same state during GRPO. The remaining GRPO implementation details are summarized in the Supplementary Material, Sec. 2.2.

This objective changes the role of feedback from an input string to a source of credit assignment. A question is rewarded only when its answer changes the retrieval state in a useful direction. Consequently, the policy can move away from feedback-covered attributes, allocate later questions to unresolved regions, and preserve evidence that remains useful after re-ranking. Because the retriever and answerer remain fixed, the ranking change directly measures question utility; the resulting policy complements accumulated memory rather than restating it, while DAR represents the evolving evidence for retrieval.

\section{Experiments}
\label{sec:experiments}

This section evaluates VFI-ReID on the VFI-PR task and analyzes its retrieval performance and learned behavior. Section 5.1 describes the models and experimental configuration. Section 5.2 compares VFI-ReID with existing interactive retrieval methods. Section 5.3 studies the contributions of the retriever, questioner, and policy optimization. Section 5.4 reports a scaling study over the number of RL states used to optimize the questioner.

\subsection{Experimental setup}
For the retriever, we adopt CLIP as the image--text backbone \cite{radford2021clip} and train it with the IRRA framework \cite{jiang2023cross} using fine-grained descriptions. We then adapt the retriever to accumulated VFI-PR evidence with DAR. For the questioner, we build on LLaVA-OneVision-Qwen2-7B-ov \cite{li2024llavaonevision} and fine-tune it with QLoRA \cite{dettmers2023qlora}, followed by retrieval-aware policy optimization. For question answering, we use Qwen2.5-7B-Instruct \cite{yang2024qwen25} to respond from the witness memory. Separately, Qwen3-VL-8B \cite{bai2025qwen3vl} generates visual-difference feedback from the same memory and the selected candidate image; it is used both to construct VFI-PEDES and to produce feedback during evaluation. Neither model receives the target image. We set the candidate-set size to $k=4$ and the interaction budget to $T=2$, following LLaVA-ReID \cite{lu2025llavareid} and MInter-PR \cite{bai2026interactive}, respectively. Further module structures and training details are provided in the Supplementary Material, Secs. 2.1 and 2.2.

\begin{figure}[t]
\centering
\includegraphics[width=\linewidth]{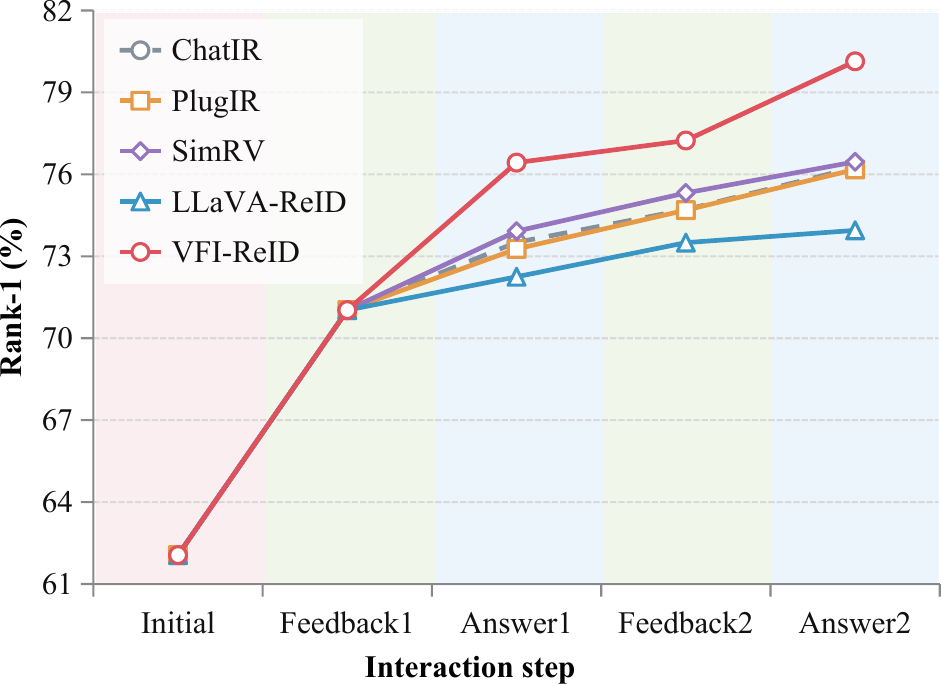}
\caption{Rank-1 progression as evidence accumulates.}
\label{fig:rank1_progression}
\end{figure}
\subsection{Comparison with State-of-the-Arts}
\label{sec:main-results}

We compare VFI-ReID with four interactive retrieval methods. ChatIR \cite{levy2023chatting} uses multiple dialogue examples to guide a pretrained language model in predicting the next question for conversational image retrieval. PlugIR \cite{lee2024plugir} selects candidates with clustering, conditions question generation on their textual descriptions, and filters redundant questions before querying the retriever. SimRV \cite{liang2023simple} extracts objects from the description and asks heuristic questions about those objects, whereas LLaVA-ReID \cite{lu2025llavareid} uses a multimodal questioner and retrieval-based supervision to learn an offline question order. Table~\ref{tab:legacy} summarizes the comparison, while Figure~\ref{fig:rank1_progression} shows the finer-grained progression across evidence states. VFI-ReID improves from 62.05 initial R@1 to 80.14 after two interaction rounds. Consistent with this improvement, its BRI is 0.385, the lowest among the compared methods, indicating that the improvement arrives earlier in the interaction. The gains occur at both visual-feedback and answer states rather than being deferred to the endpoint: visual feedback exposes a discrepancy that is not present in the current text, the questioner seeks another unresolved attribute, and the answer turns that attribute into retrieval evidence. Together, these updates let VFI-ReID convert alternating evidence acquisition and re-ranking into progressively stronger retrieval quality.

\begin{figure}[H]
\centering
\includegraphics[width=\linewidth]{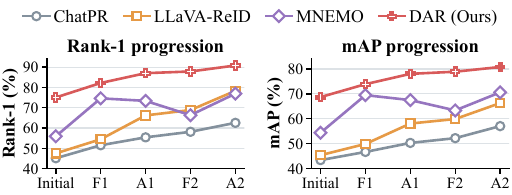}
\caption{Retrieval progression on the common CUHK-PEDES subset. F1 and F2 denote the first and second visual-feedback states, while A1 and A2 denote the corresponding answer states.}
\label{fig:cuhk-progress}
\end{figure}

In addition, to compare how different retrieval systems use accumulating evidence under a common data source, we evaluate ChatPR, LLaVA-ReID, MNEMO, and DAR on the CUHK-PEDES portion of VFI-PEDES. Because their underlying data sources overlap with CUHK-PEDES, we select the 2,360 VFI-PEDES test trajectories whose target images originate from the CUHK-PEDES subset used to construct VFI-PEDES and retain the same gallery and five cumulative evidence states for every method: the initial description, first feedback, first answer, second feedback, and second answer. Figure~\ref{fig:cuhk-progress} reports the initial retrieval levels and traces R@1 and mAP as evidence accumulates. ChatPR and LLaVA-ReID start at similar levels, whereas MNEMO obtains a large first-feedback gain but fluctuates at later states. DAR starts substantially higher and improves consistently through both feedback and answer updates, reaching the strongest final retrieval quality.

\begin{table}[t]
\centering
\caption{Component ablation for VFI-ReID. VF denotes visual feedback.}
\label{tab:ablation}
\small
\begin{tabular*}{\columnwidth}{@{\extracolsep{\fill}}cccc|cccc@{}}
\toprule
VF & DAR & SFT & RL & R@1 & R@5 & R@10 & mAP \\
\midrule
\checkmark &  &  &  & 46.79 & 70.07 & 79.44 & 33.72 \\
\checkmark &  & \checkmark & \checkmark & 65.25 & 85.20 & 90.71 & 43.84 \\
\checkmark & \checkmark &  &  & 72.90 & 90.34 & 94.67 & 54.22 \\
\checkmark & \checkmark & \checkmark &  & 73.96 & 90.45 & 94.85 & 54.65 \\
 & \checkmark & \checkmark & \checkmark & \underline{75.22} & \underline{91.54} & \underline{95.38} & \underline{54.68} \\
\textbf{\checkmark} & \textbf{\checkmark} & \textbf{\checkmark} & \textbf{\checkmark} & \textbf{80.14} & \textbf{93.80} & \textbf{96.66} & \textbf{57.77} \\
\bottomrule
\end{tabular*}
\end{table}

\subsection{Ablation study}
\label{sec:ablation}

Table~\ref{tab:ablation} compares six configurations to isolate the fine-grained-description retriever, DAR, SFT, RL, and visual feedback: the bare model, SFT+RL without DAR, DAR alone, DAR+SFT, DAR+SFT+RL without visual feedback, and the full model. Rows without DAR retain the fine-grained-description retriever, whereas the rows with DAR use the retriever adapted to accumulated evidence. This setup separates the effect of changing the retrieval state from the effect of training the questioner on that state, while keeping the evaluation procedure fixed. The ablation results show that the full combination is strongest across all retrieval metrics. The bare model reaches 46.79 R@1, while adding SFT and RL without DAR raises it to 65.25, showing that questioner training is useful even before retriever adaptation. DAR alone reaches 72.90 R@1, and adding SFT raises it to 73.96 by providing a stable questioner for the new evidence format. With DAR, SFT, and RL fixed, removing visual feedback lowers the second-round result to 75.22 R@1 and 54.68 mAP, whereas enabling it raises the result to 80.14 R@1 and 57.77 mAP. RL and feedback therefore provide complementary gains: policy optimization aligns question selection with retrieval consequences, while visual feedback supplies candidate-conditioned evidence.

\subsection{Scaling study}
\label{sec:scaling}

We study the trade-off among retrieval accuracy, interaction efficiency, and optimization cost as the RL state pool grows. The scaling variable is the number of interaction states used to optimize the questioner, rather than the number of question types or question--answer pairs. Each state contains accumulated evidence, a candidate set, and the retrieval outcome used for policy optimization. We compare 0K, 1K, 3K, and 5K states; 0K disables the RL stage and corresponds to DAR+SFT. Table~\ref{tab:scaling} makes this three-way trade-off explicit: R@1 rises from 73.96 with no RL states to 79.53 with 1K and 80.14 with 3K, while the 5K setting provides only a modest further increase to 80.29. BRI improves from 0.445 to 0.397 and then 0.385 as the pool grows to 3K, but slightly worsens to 0.391 at 5K, indicating that the 3K policy reaches useful rankings earlier in the interaction. Thus, 5K offers the highest endpoint accuracy but incurs additional construction and optimization cost, whereas 3K retains nearly the same accuracy with better interaction efficiency and lower cost; we use 3K for the main model.

\begin{table}[t]
\centering
\caption{Scaling study with different numbers of RL states.}
\label{tab:scaling}
\small
\begin{tabular*}{\columnwidth}{@{\extracolsep{\fill}}c|ccccc@{}}
\toprule
RL states & R@1 & R@5 & R@10 & mAP & BRI$\downarrow$ \\
\midrule
0K & 73.96 & 90.45 & 94.85 & 54.65 & 0.445 \\
1K & 79.53 & 93.73 & 96.61 & 57.62 & 0.397 \\
3K & \underline{80.14} & \textbf{93.80} & \underline{96.66} & \underline{57.77} & \textbf{0.385} \\
5K & \textbf{80.29} & \underline{93.77} & \textbf{96.84} & \textbf{58.45} & \underline{0.391} \\
\bottomrule
\end{tabular*}
\end{table}

\section{Conclusion}
\label{sec:conclusion}

We presented VFI-ReID, a framework for VFI-PR that closes the loop between retrieved candidates, visual feedback, question selection, and cross-modal re-ranking. The VFI-PEDES dataset records visual-difference feedback and accumulated evidence states for reproducible training and evaluation. DAR adapts the retriever to these dynamic states, while supervised warm-up and retrieval-aware RL improve the question policy as interaction proceeds. The resulting dataset and modular framework make it possible to study whether a question is useful because it sounds discriminative or because its answer improves retrieval. Future work will investigate longer conversations and uncertainty-aware policies that decide when visual-difference feedback is reliable enough to enter retrieval memory.

{\small
\bibliographystyle{unsrtnat}
\bibliography{main}
}

\end{document}